\documentclass[6pt]{article}
\usepackage[utf8]{inputenc}
\usepackage[verbose=true,letterpaper]{geometry}
\AtBeginDocument{
  \newgeometry{
    textheight=9in,
    textwidth=6.5in,
    top=1in,
    headheight=14pt,
    headsep=25pt,
    footskip=30pt
  }
}

\usepackage{arxiv}
 \usepackage[table,xcdraw]{xcolor}
\usepackage{graphicx}
\usepackage{amsmath,amssymb}
\usepackage{algorithm}
\usepackage{algorithmic}
\usepackage{pslatex}
\usepackage{apacite}
\usepackage{soul}
\usepackage{subfig}
\usepackage{float}
\usepackage{csquotes}
\usepackage{balance}
\usepackage{subfig}

\newcommand{\model}{S}

\title{Cultural evolution via iterated learning and communication \\ explains efficient color naming systems}

\author{\textbf{Emil Carlsson (caremil@chalmers.se)} \\ 
  \textbf{Devdatt Dubhashi (dubhashi@chalmers.se)} \\ 
  Department of Computer Science and Engineering, Chalmers University of Technology, Gothenburg, Sweden. \\
  \\
 \textbf{Terry Regier (terry.regier@berkeley.edu)} \\ 
 Department of Linguistics, UC Berkeley, Berkeley, CA, USA}
\date{}
\begin{document}

\maketitle

\begin{abstract}
It has been argued that semantic systems reflect pressure for efficiency, and a current debate concerns the cultural evolutionary process that produces this pattern.  We consider efficiency as instantiated in the Information Bottleneck (IB) principle, and a model of cultural evolution that combines iterated learning and communication.  We show that this model, instantiated in neural networks, converges to color naming systems that are efficient in the IB sense and similar to human color naming systems.  We also show that some other proposals such as iterated learning alone,  communication alone, or the greater learnability of convex categories, do not yield the same outcome as clearly.  We conclude that the combination of iterated learning and communication provides a plausible means by which human semantic systems become efficient.


\end{abstract}

\section{Introduction}

Semantic categories vary across languages, and it has been proposed that this variation can be explained by  functional pressure for efficiency.  On this view, systems of categories are under pressure to be both simple and informative (e.g.\ \citeNP{rosch-1978}), and different languages arrive at different ways of solving this problem, yielding wide yet constrained cross-language variation.  There is evidence for this view from semantic domains such as kinship \cite{Kemp2012}, container names \shortcite{xu-et-al-2016}, names for seasons \shortcite{kemp-et-al-2019}, indefinite pronouns \shortcite{denic-et-al-CSj-2022}, modals \shortcite{imel-sst-SALT-2022}, and numeral systems (\shortciteNP{Xu2020}, and relatedly \shortciteNP{denic-szymanik-CSj-2024}).  \shortciteA{Zaslavsky2018a} gave this proposal a firm theoretical foundation by grounding it in an independent information-theoretic principle of efficiency, the Information Bottleneck
(IB) principle \shortcite{Tishby-1999}; they also showed: (1) that color naming systems across languages are efficient in the IB sense, (2) that optimally IB-efficient systems resemble those found in human languages, and (3) that the IB principle accounts for important aspects of the data that had eluded earlier explanations.  
Subsequent work has shown that container naming \shortcite{Zaslavsky2019c}, grammatical categories of number, tense, and evidentiality \shortcite{Mollica21},  and person systems \shortcite{Zaslavsky2021} are also efficient in the IB sense.

In a commentary on this line of research, \citeA{levinson-kinship-2012} asked how semantic systems evolve to become efficient, and suggested that an important role may be played by iterated learning (IL; e.g.\ \shortciteNP{scott-phillips-kirby-2010}).  In IL, a cultural convention is learned by one generation of agents, who then provide training data from which the next generation learns, and so on.  The convention changes as it passes through generations, yielding a cultural evolutionary process.  The idea that such a process could eventually lead to efficient semantic systems has since been explored and broadly supported.  \shortciteA{xu-et-al-2013} showed that chains of human learners who were originally given a randomly generated color category system eventually produced systems that were similar to those of the World Color Survey (WCS; \shortciteNP{cook-2005}), a large dataset of color naming systems from 110 unwritten languages.  Although this study did not directly address efficiency, 
\shortciteA{Carstensen15} drew that link explicitly: they reanalzyed the data of \shortciteA{xu-et-al-2013} and showed that the color naming systems produced by IL not only became more similar to those of human languages -- they also became more informative; the same paper also presented analogous findings for semantic systems of spatial relations.  In response, \shortciteA{Carr20} argued, on the basis of a Bayesian model of IL and experiments with human participants, that learning actually contributes simplicity rather than informativeness.  Overall, there is support for the idea that IL can lead to efficient semantic systems, with continuing debate over how and why.  
There are also recent proposals that non-iterated learning -- e.g.\ in the context of a dyad of communicating agents (e.g.\ \shortciteNP{Kageback2020,Chaabouni,tucker-et-al-2022}), or in a single agent without communication (e.g.\ \shortciteNP{Steinert-Threlkeld20,Gyevnar22}) -- can explain efficient color naming systems.  In particular, \shortciteA{Steinert-Threlkeld20} argued that ``[e]ase of learning explains semantic universals'' (see also \shortciteNP{genter-bowerman-2009}).  To illustrate this point,  \shortciteA{Steinert-Threlkeld20} highlighted the greater learnability, in a neural network, of convex as opposed to non-convex color categories, in line with earlier proposals arguing for the importance of convexity in conceptual space as an important constraint on human semantic categories (\shortciteNP{Gardenfors00, Jager2007,Jager10}).  These recent contributions build on an important line of earlier work 
using agent-based simulations cast as evolutionary models, without explicitly addressing efficiency
(e.g.\ \shortciteNP{Steels05, belpaeme-bleys-2005, Dowman07, 
Jameson09, Baronchelli10}). 


Several of these prior studies have engaged efficiency in the IB sense, and two are of particular relevance to our own work.  \shortciteA{Chaabouni} showed that a dyad of neural network agents, trained to discriminate colors via communication, eventually arrived at color naming systems that were highly efficient in the IB sense.  However, these systems did not always resemble those of human languages: their categories ``depart to some extent from those typically defined by human color naming'' (\shortciteNP{Chaabouni}, p. 11 of SI).  \shortciteA{tucker-et-al-2022} explored a similar color communication game, and found that their neural agents gravitated to color naming systems that are both essentially optimally efficient in the IB sense, and similar to human color naming systems from the WCS.  They achieved this by optimizing an objective function that is based on the IB objective.  To our knowledge, earlier work leaves open whether both high IB efficiency and similarity to human languages can be achieved through processes and principles that are independent of IB.  We explore that question here.  We also wish to establish here whether  such independent principles may address the one case in which IB-optimal color naming systems deviate to some extent from empirical observation: the case of 3-term systems \shortcite[p.\ 7941]{Zaslavsky2018a}.  Overall, we wished to ascertain whether a natural model of cultural evolution might account both for the many cases in which IB matches the data, and for the one case in which it deviates from the data to some extent.


In what follows, we first 
demonstrate that there exist many possible color naming systems that are highly efficient in the IB sense, but do not closely resemble human systems.
The fact that there exist such efficient-yet-not-human-like systems is not surprising given that IB is a non-convex optimization problem \shortcite{Tishby-1999,Zaslavsky2018a}, but appreciating the prevalence of such systems may be helpful in understanding how \shortciteA{Chaabouni} achieved high IB efficiency with systems that deviate from human ones.  We then show that iterated learning, instantiated in communicating neural networks, gravitates toward efficiency and, within the class of efficient systems, gravitates more toward human color naming systems than toward others.  Finally, we show that iterated learning alone, communication alone, and convexity alone, do not yield that outcome as clearly.
We conclude that iterated learning and communication jointly provide a plausible explanation of how human color naming systems become efficient.

\section{Not all efficient systems are human-like}\label{sec:not_all}

\begin{figure}
    \centering
    \includegraphics[width=0.8\textwidth]{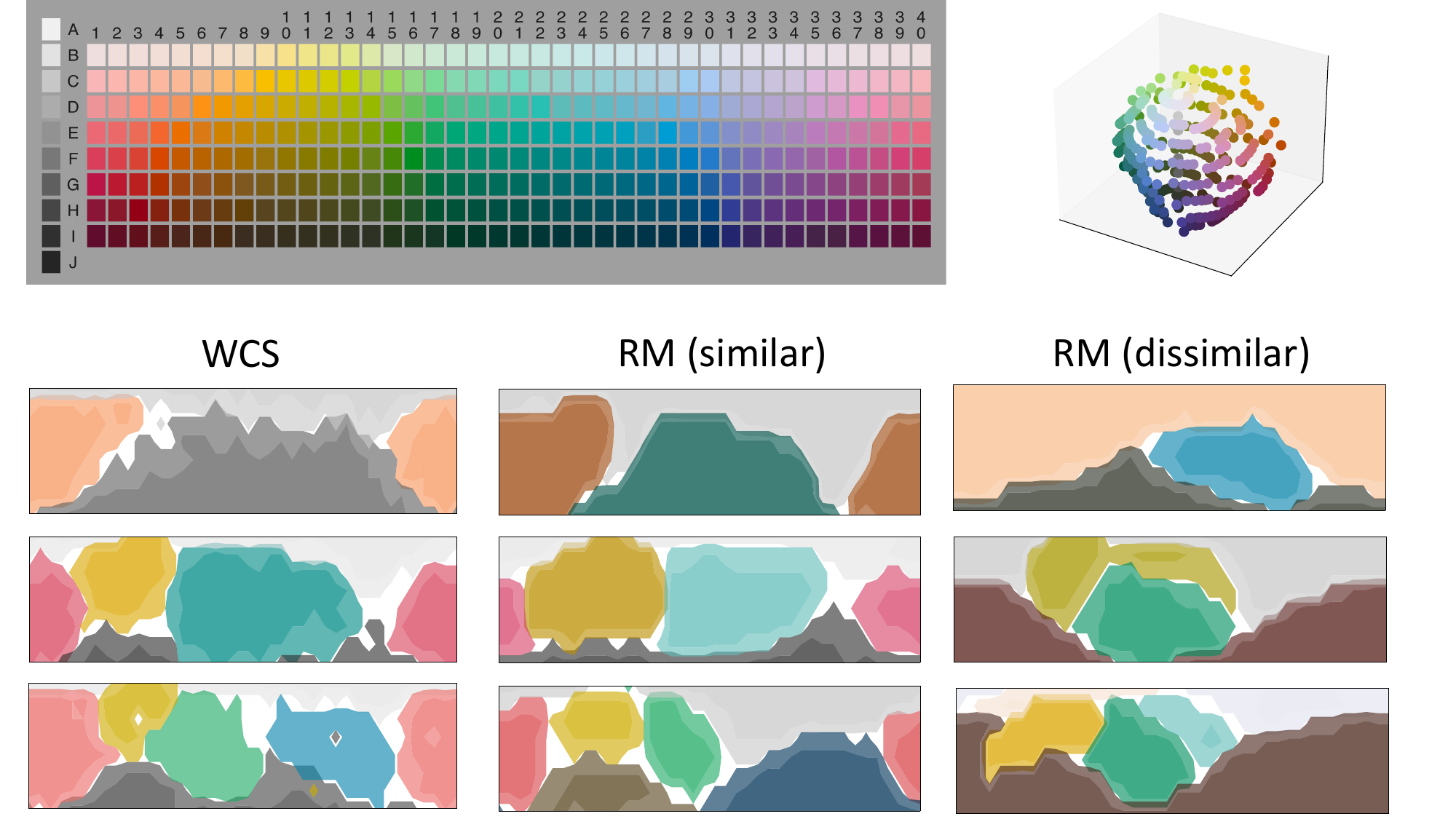}

    \caption{Top: Color naming stimulus grid (left), and  stimuli plotted in CIELAB space (right). Bottom: 9 color naming systems displayed relative to the grid.  The left column contains color naming systems from 3 languages in the WCS. Colored regions indicate category extensions, and the color code used for each category is the mean  of that category in CIELAB color space.  The named color categories are distributions, and for each category we highlight the level sets between $0.75-1.0$ (unfaded area) and $0.3-0.75$ (faded area).  The middle and right columns contain randomly-generated systems of complexity comparable to that of the WCS system in the same row.  The middle column shows random systems that are similar to the WCS system in the same row.  The right column shows random systems that are dissimilar to the WCS system in the same row; at the same time, there is no other WCS system that is more similar to this random system.}
    \label{fig:maps}
\end{figure}

We considered a natural class of artificial color naming systems (see e.g.\ \shortciteNP{Abbott2016FocalCA,zaslavsky-et-al-2022-nafaanra}). 
In this class, each named category $w$ is modeled as a spherical Gaussian-shaped kernel with mean (prototype) $x_w$ in 3-dimensional CIELAB color space (Figure \ref{fig:maps}, top right panel), such that the distribution over words $w$ given a color chip $c$ at location $x_c$ in CIELAB space is: 
\begin{align}\label{eq:random}
    \model(w|c) \propto e^{- \eta||x_c - x_w||_2^2}
\end{align}
where $\eta > 0$ is a parameter controlling the precision of the Gaussian kernel.  We then generated artificial color category systems with $K=3 \dots 10$ categories each, by first sampling $\eta$ randomly from a uniform distribution over the interval $[0.001, 0.005]$ for each system  and then sampling the prototype $x_w$ of each category $w$ randomly, without replacement, from a uniform distribution over the cells of the color naming grid shown in the top left panel of Figure \ref{fig:maps}; this shows the same set of colors as in the top right panel, but now in a 2-D array.  In analyzing these systems, we draw on four quantities from the IB framework as presented by \shortciteA{Zaslavsky2018a} and reviewed below in Appendix~\ref{appendix:IB}: the complexity of a category system, the accuracy of a category system, $\epsilon$ (a measure of the inefficiency of a category system, or its deviation from the theoretical limit of efficiency), and gNID (a measure of dissimilarity between two category systems).  We noted that the range of complexity (in the IB sense) for systems in the World Color Survey (WCS) was $[0.84, 2.65]$, and also noted that our random model sometimes generated systems outside this range; we only considered artificial systems with complexity within this range, and generated $100$ such systems for each $K$; we refer to these systems as RM, for random model.  

The lower panels of Figure \ref{fig:maps} compare natural color naming systems to artificial RM systems.  The leftmost column shows three attested color naming systems from the WCS, from top to bottom: B\'et\'e (iso: bev, C\^ote d'Ivoire), Colorado / Tsafiki (iso: cof, Ecuador), and Dyimini (iso: dyi, C\^ote d'Ivoire).  The middle column shows RM systems that are similar to the WCS system in the same row, and the rightmost column shows RM systems that are dissimilar to the WCS system in the same row but of about the same complexity. In each row, the rightmost system, which is dissimilar to the WCS system in that row, is nonetheless more similar to that WCS system than to any other WCS system;  this means that it is dissimilar to all WCS systems.  Thus, there exist RM systems that are quite similar to naturally occurring systems, and other RM systems that are quite dissimilar to naturally occurring systems.  To quantify this pattern, we separated the RM systems into two groups, based on whether their gNID to the closest WCS system exceeded a threshold.  
We set this threshold to the smallest gNID between systems in the left (WCS) and right (RM dissimilar) columns of Figure \ref{fig:maps}, which is $0.29$.  We then grouped all RM systems with gNID to the closest WCS system below this threshold into one group, $\text{RM}_{\text{s}}$ (for similar to WCS), and the other RM systems into another group, $\text{RM}_{\text{d}}$ (for dissimilar to WCS).  We found that $38\%$ of the RM systems fell in $\text{RM}_{\text{d}}$ and they spanned the complexity range $[0.86, 2.26]$.   Thus, a substantial proportion of the RM systems are at least as dissimilar to WCS systems as are those in the right column of Figure \ref{fig:maps}.

\begin{figure}[h]
    \centering
    \includegraphics[width=0.8\textwidth]{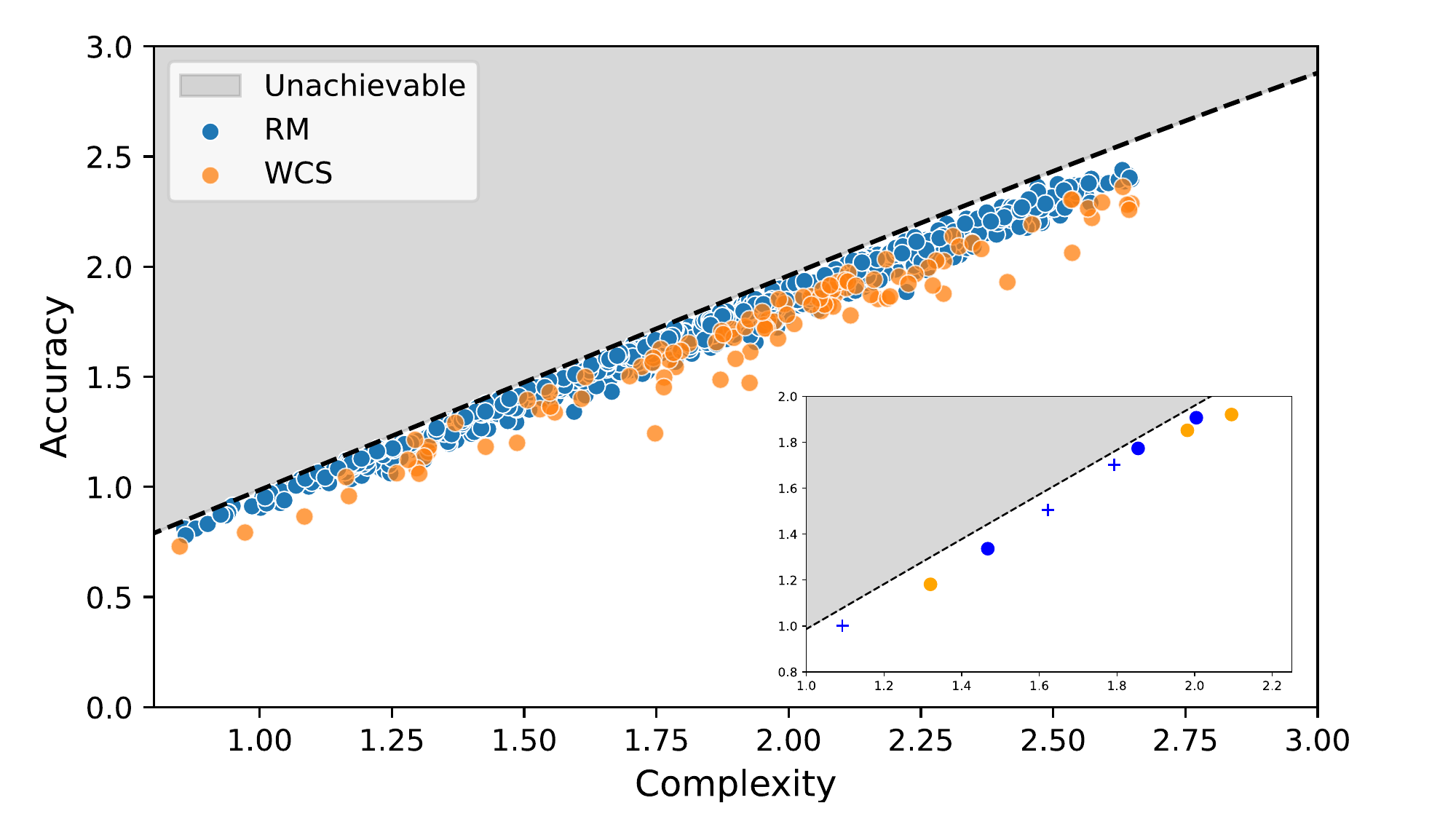}
    \caption{Efficiency of color naming, following Zaslavsky et al., 2018.  The dashed line is the IB theoretical limit of efficiency for color naming, indicating the greatest possible accuracy for each level of complexity.  The color naming systems of the WCS are shown in orange, replicating the findings of Zaslavsky et al., 2018.  Our RM systems are shown in blue.  It can be seen that the RM systems are often closer to the IB curve than the WCS systems are.  The inset shows the 9 color systems of Figure \ref{fig:maps}, with the dissimilar random systems shown as +.}
    \label{fig:ib}
\end{figure}

Figure \ref{fig:ib} shows the results of an IB efficiency analysis of the WCS systems (replicating \shortciteNP{Zaslavsky2018a}, and assuming their least-informative prior), and also of our RM systems.  It can be seen that all RM systems are highly efficient in the IB sense -- i.e.\ they are close to the IB curve that defines the theoretical limit of efficiency in this domain.  Mann-Whitney $U$ tests revealed (1) that the RM systems tend to exhibit greater efficiency (lower inefficiency $\epsilon$) than do the WCS systems in the same complexity range 
($P \ll .001$), and (2) that the $\text{RM}_{\text{d}}$ systems, which are dissimilar to WCS systems, are also more efficient than WCS systems ($P \ll .001$, one-sided), and slightly to marginally more efficient than $\text{RM}_{\text{s}}$ systems ($P =.019$ one-sided; Bonferroni corrections do not change the qualitative outcome). 
These findings suggest that there is a substantial number of color naming systems that are dissimilar to those of human languages, yet more efficient than them.  This in turn may help to make sense of 
\shortciteauthor{Chaabouni}'s 
\citeyear{Chaabouni} 
finding that their evolutionary process yielded systems that were highly efficient but not particularly similar to human ones: our analysis illustrates that there are many such systems.  Given this, we sought an evolutionary process that would yield both efficiency in the IB sense, and similarity to human systems, grounded in processes and principles independent of IB (cf. \shortciteNP{tucker-et-al-2022}).

\section{Iterated learning and communication}

As noted above, iterated learning (IL; e.g.\ \shortciteNP{Kirby01, Smith03}) is a cultural evolutionary process in which a cultural convention is learned first by one generation of agents, who then pass that convention on to another generation, and so on --- and the convention changes during inter-generational transmission. Some of the work we have reviewed above addresses IL 
(e.g.\ \shortciteNP{levinson-kinship-2012,Carstensen15,Carr20}).  However other work we have reviewed instead addresses cultural evolution through communication within a single generation (e.g.\ \shortciteNP{Kageback2020,Chaabouni,tucker-et-al-2022}).  We wished to explore the roles of both IL and communication, and so we adopted an approach that involves both, in a way that allows the role of each to be highlighted.  Specifically, we adopted the recently proposed \emph{neural iterated learning} (NIL) algorithm \shortcite{Ren2020Compositional}. In the NIL algorithm, artificial agents are implemented as neural networks that communicate with each other within a generation, and cultural convention 
(in our case, a color naming system) 
evolves both from within-generation communication and from inter-generational transmission, as the convention is iteratively passed down through generations of artificial agents, with each new generation learning from the previous one.\footnote{NIL, or neural iterated learning, is therefore not an entirely informative name for this process, as it does not explicitly label the important element of within-generation communication.}

\begin{figure}[h]
    \centering
    \includegraphics[width=0.7\textwidth]{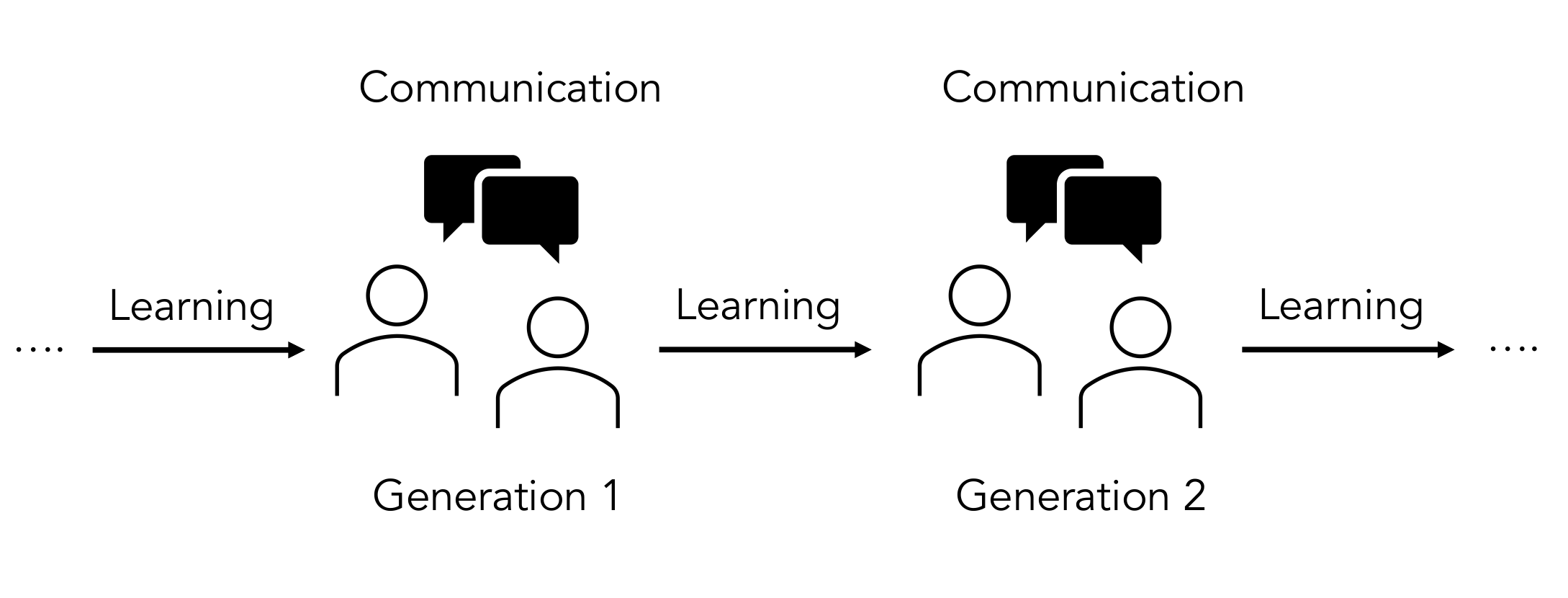}
    \caption{Illustration of the neural iterated learning (NIL) algorithm (Ren et al., 2020). The algorithm alternates between communication within a generation, and learning that is iterated across generations.}
    \label{fig:enter-label}
\end{figure}

In the NIL algorithm, each generation $t$ (for time step) consists of two artificial agents, a speaker $S_t$ and a listener $L_t$. The NIL algorithm operates in three phases. (1) In the first phase, the \emph{learning phase}, both agents are exposed to the naming convention of the previous generation. This is done by first training the speaker $S_t$, using cross-entropy loss, on color-name pairs generated by the speaker of the previous generation. The listener $L_t$ is then trained via reinforcement learning in a few rounds of a signaling game while keeping $S_t$ fixed: that is, the speaker learns from the previous generation, and the listener then learns from the speaker. We had the agents play the signaling game used by \shortciteNP{Kageback2020}, in which the speaker is given a color chip $c$, sampled from a prior distribution over color chips, and produces a category name describing that color. The listener then attempts to identify the speaker's intended color based on the name produced, by selecting a color chip $\hat{c}$ from among those of the naming grid shown in Figure \ref{fig:maps}.  A reward is given to the listener depending on how perceptually similar the selected chip is to the original color, following Equation~\ref{eqn:reward} below. 
(2) In the second phase, the \emph{interaction phase}, the agents play the same signaling game but this time both agents receive a joint reward and update their parameters during communicative interactions. (3) In the third phase, the \emph{transmission phase}, color-name pairs are generated by sampling colors from the prior distribution and obtaining names for them from the speaker $S_t$. These color-name pairs are then passed on to the next generation of agents. In all three phases, color chips are sampled according to the least-informative prior of \citeA{Zaslavsky2018a}. Algorithm \ref{alg:NIL} presents a schematic overview of the NIL algorithm, and \citeA{Ren2020Compositional} present a detailed description. 

For the main experiments we represent both the speaker and listener as neural networks with one hidden layer consisting of $25$ units with a sigmoidal activation function.  Individual colors are represented in 3-dimensional CIELAB space when supplied as input to the speaker, and category names as one-hot encoded vectors.  For the reinforcement learning parts of NIL we use the classical algorithm REINFORCE \shortcite{Williams92}. For the transmission phase we sample $300$ color-name pairs with replacement, out of the $330$ chips in the entire stimulus set; this ensures that the new generation will have seen examples from most of color space but it is impossible for them to have seen all color-name pairs.  To optimize the neural networks, we  use the optimizer Adam \shortcite{Kingma2014}, both in the learning and interaction phase, with learning rate $0.005$ and batch size $50$. For each phase in the NIL algorithm we take $1000$ gradient steps. We stop the NIL algorithm either after $250$ generations or once the maximum difference in IB complexity and accuracy over the ten latest generations is smaller than $0.1$ bit, i.e.\ when the last ten generations are all within a small region of the IB plane.

\begin{algorithm}[h]
\begin{algorithmic}[1]
\STATE Initialize dataset $D_1$ uniformly at random
\FOR{$t=1 ... $}
\STATE {\bf\emph{Learning Phase}}
\STATE Randomly initialize $S_t$ and $L_t$.
\STATE Train $S_t$ on $D_t$ using stochastic gradient descent and cross-entropy loss.
\STATE Play signaling game between $S_t$ and $L_t$ and update parameters of only $L_t$ using the rewards.
\STATE {\bf\emph{Interaction Phase}}
\STATE Play signaling game between $S_t$ and $L_t$ and update parameters of {\bf both} agents using the rewards.
\STATE {\bf \emph{Transmission Phase}}
\STATE Create transmission dataset $D_{t+1}$ consisting of color-name pairs, $(c, w)$ by sampling colors from the prior $p(c)$ and providing them as input to $S_t$.
\ENDFOR
\end{algorithmic}
\caption{Neural Iterated Learning}\label{alg:NIL}
\end{algorithm}

\paragraph{The reward function:}
The reward function of \shortciteNP{Kageback2020}, which we use here, takes the form: 
\begin{align}
    r(c, \hat{c}) = e^{-\gamma||x_c-x_{\hat{c}}||_2^2}
    \label{eqn:reward}
\end{align}
where $c$ is the chip sampled by the speaker, $\hat{c}$ is the chip chosen by the listener as their interpretation of the chip intended by the speaker, $x_c$ is the location in CIELAB space of chip $c$, and $\gamma$ is a parameter that controls how precise the listener's choice $\hat{c}$ has to be. As $\gamma \rightarrow \infty$ the above reduces to a binary reward function, i.e.\ the listener has to perfectly reconstruct the color to get any reward. On the other hand, if $\gamma=0$ the reward function is vacuous in the sense that any possible reconstruction yields a reward of $1$. We use $\gamma=0.001$ which was originally used by \shortciteNP{Kageback2020} and motivated by the analysis in \shortciteNP{Regier2007}. 


\section{Analyses and results}

\subsection{Iterated learning and communication operating together}

\begin{figure}[h]
    \centering
    \begin{tabular}{cc}
    \multicolumn{2}{c}{\includegraphics[width=0.75\textwidth]{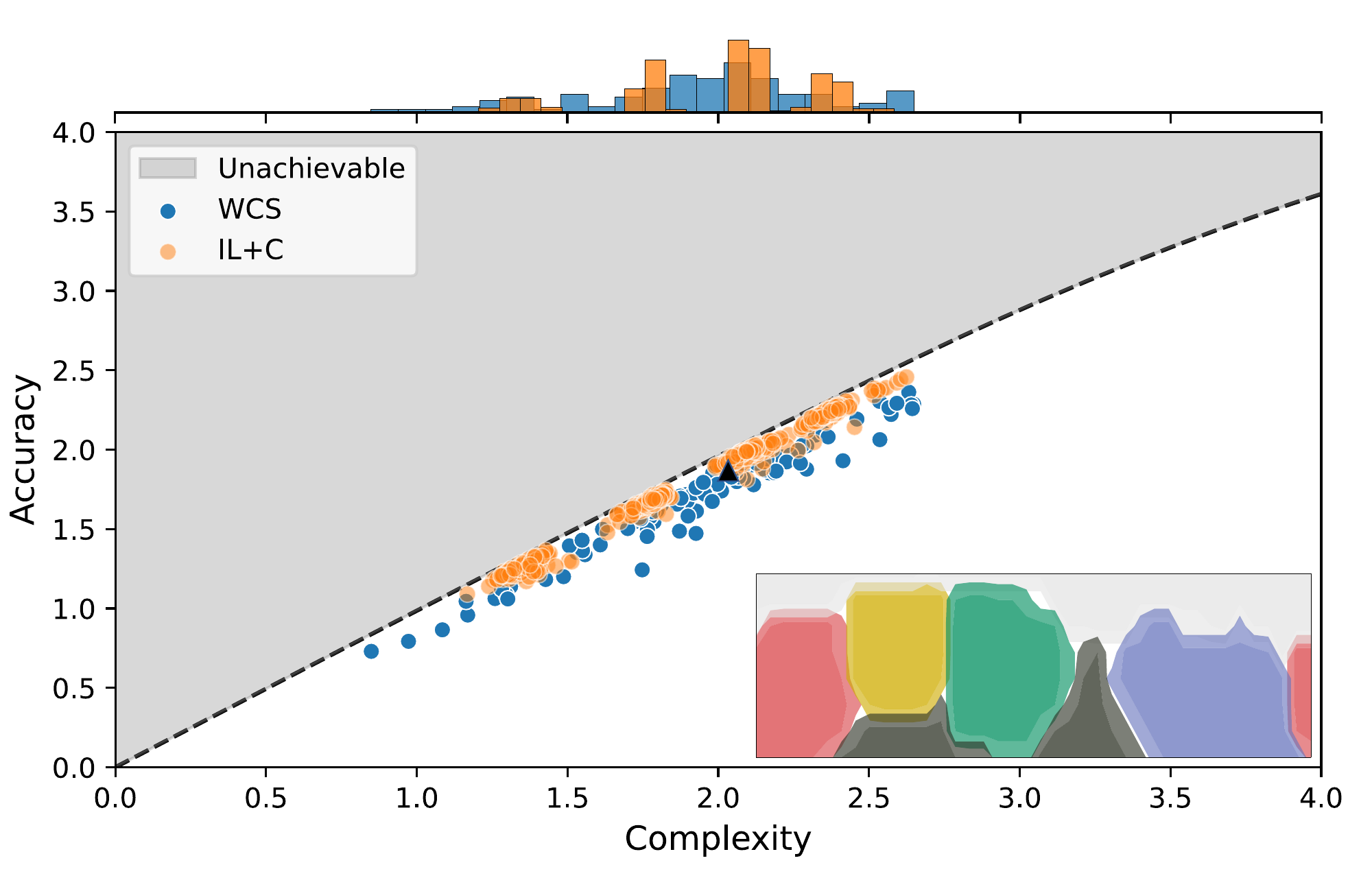} }
    \\
    \includegraphics[width=0.48\textwidth]{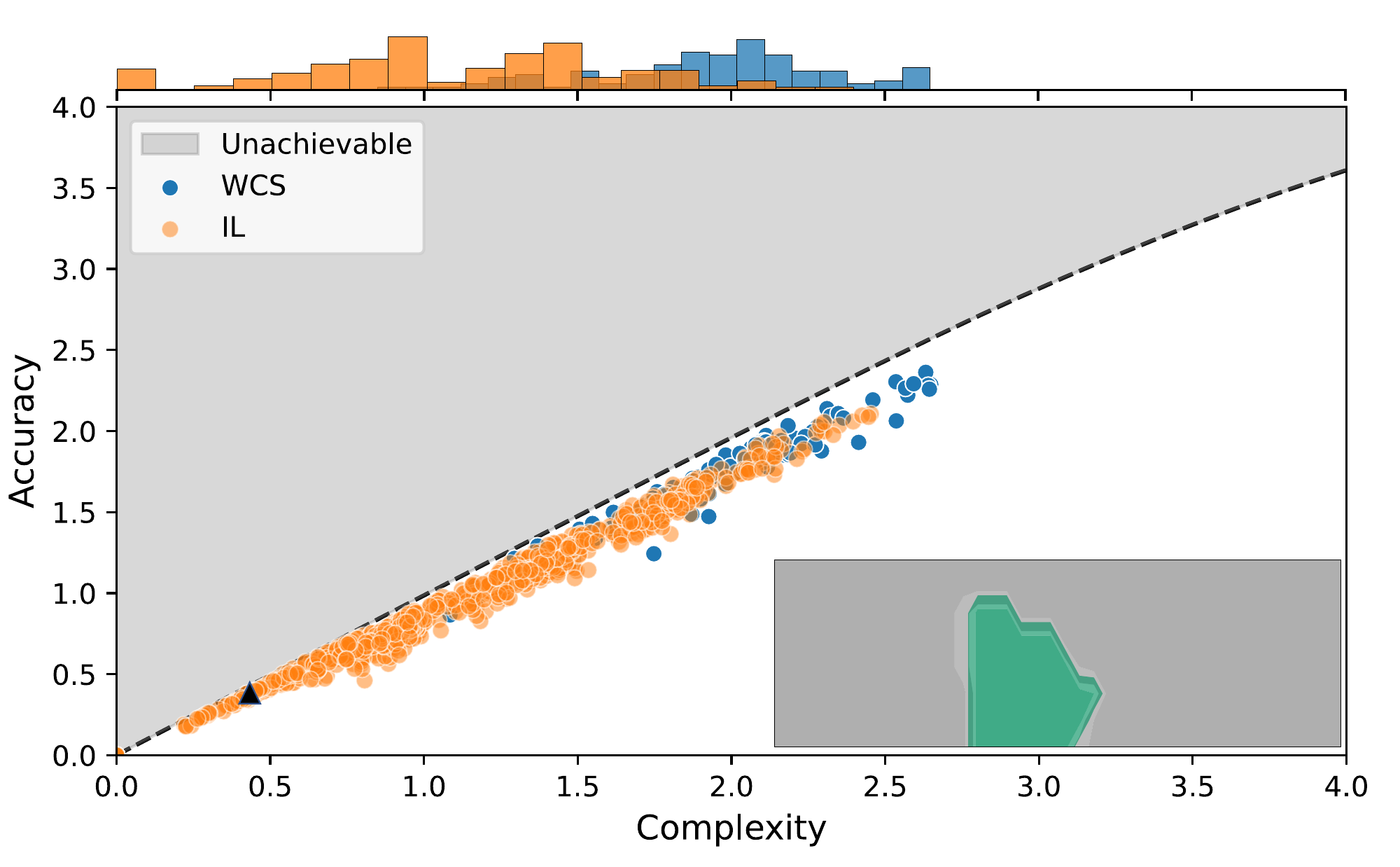} &
    \includegraphics[width=0.48\textwidth]{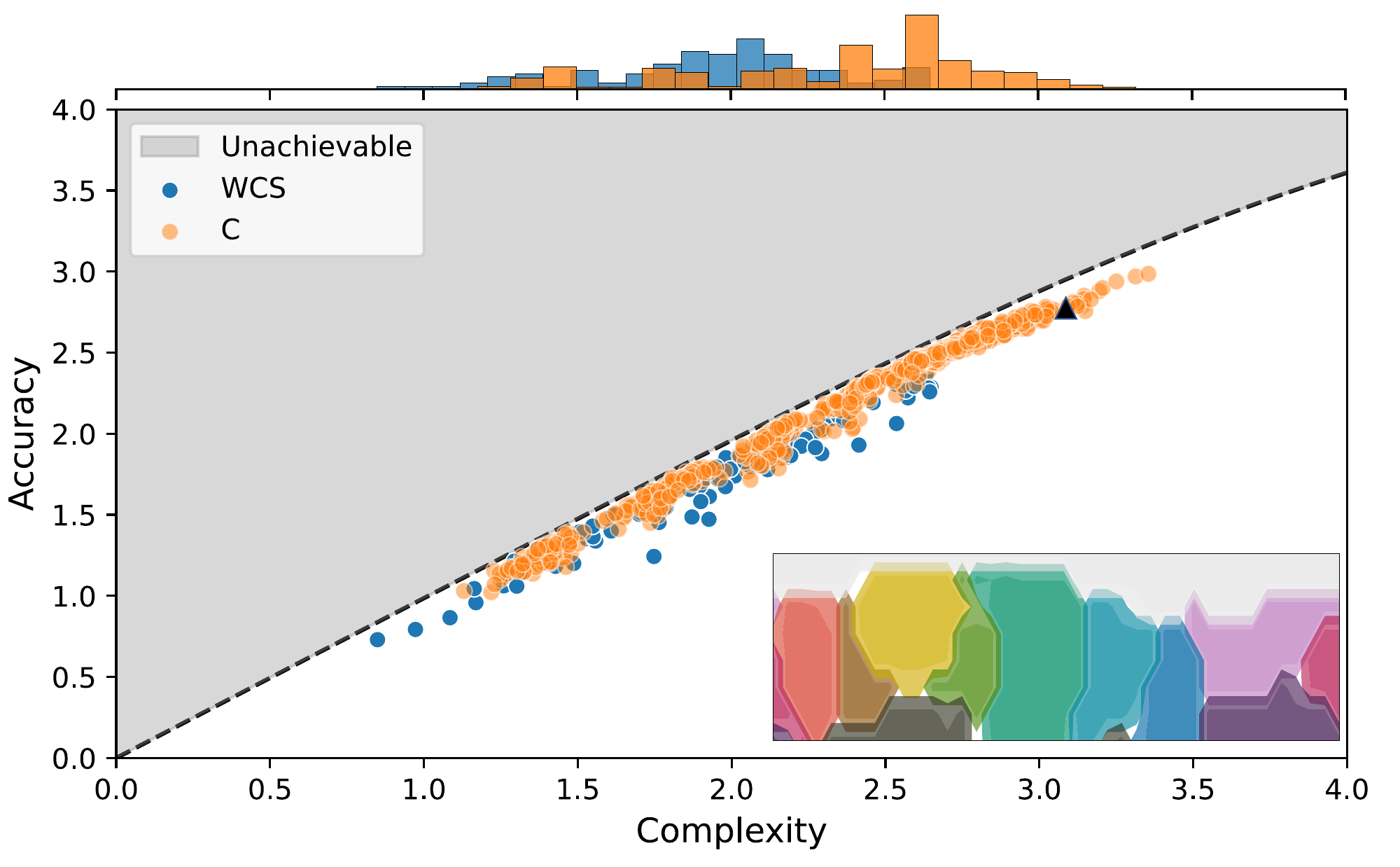} \\
    \end{tabular}
    
    \caption{Efficiency of the (top) IL+C, (bottom left) IL, and (bottom right) C evolved color naming systems (orange dots), in each case compared with the natural systems of the WCS (blue dots). The black triangle indicates the end state of one run, shown in the inset color map. The histograms above each figure indicate the proportion of systems at the corresponding complexity level. }
    \label{fig:IL-vs-WCS}
\end{figure}

For each vocabulary size $K=3 \dots 10$ and $K=100$ 
we ran $100$ independent instances of the NIL algorithm. For each instance, we considered the color naming system of the last speaker to be the result of that instance --- we call these systems IL+C, as they are the result of iterated learning plus communication, and we evaluated the IL+C systems in the IB framework. As can be seen in Figure \ref{fig:IL-vs-WCS} (top panel), the IL+C systems are highly efficient in the IB sense: they lie near the theoretical efficiency limit (median inefficiency $\epsilon = 0.07$), and they are no less efficient than the random RM systems we considered above (median inefficiency $\epsilon = 0.09$), which in turn are more efficient than the human systems of the WCS (see above).  Thus, iterated learning plus communication as formalized in the NIL algorithm leads to semantic systems that are efficient in the IB sense.  This is consistent with existing proposals: the reward during the signaling game favors informativeness (higher reward for similar colors, following \shortciteNP{Kageback2020}), and it has been argued that learning favors simplicity (e.g.\ \shortciteNP{Carr20}).  
Interestingly, all the resulting systems lie within the complexity range of the WCS systems even though NIL could theoretically produce much more complex systems, especially when initialized with $K=100$.

\begin{figure}[t]
    \centering
    \includegraphics[width=0.8\columnwidth]{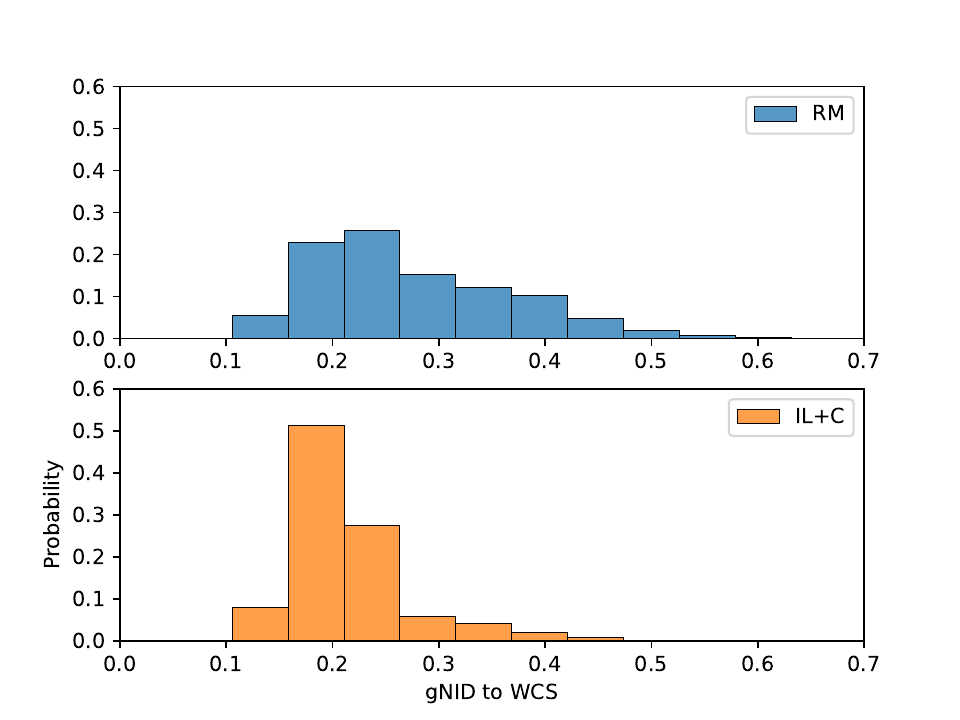}
    \caption{Distribution of dissimilarity to WCS systems (minimum gNID to any WCS system), shown for IL+C and RM systems.  The RM systems include both $\text{RM}_{\text{s}}$ and $\text{RM}_{\text{d}}$.  Evolved IL+C systems tend to be more similar to attested WCS systems than are random but highly efficient RM systems.}
    \label{fig:dist_to_wcs}
\end{figure}

\shortciteA{xu-et-al-2013} showed that chains of iterated human learners tended to gravitate toward color naming systems that were similar to those of the WCS, and we wished to know whether the same was true of computational agents in the NIL framework.  For each IL+C system, we determined the dissimilarity (gNID) between that system and the most similar (lowest gNID) WCS system.  We also determined the analogous quantity (dissimilarity to the most similar WCS system) for each random RM system.  Figure \ref{fig:dist_to_wcs} shows that IL+C systems tend to be similar to WCS systems to a greater extent than RM systems do, and this was confirmed by 
a one-sided Mann-Whitney $U$ test $(P \ll.001)$.
Thus, the NIL process tends to gravitate toward human (WCS) systems to a greater extent than a random but efficient baseline, RM.  

\begin{figure}[t]
    \centering
    \includegraphics[width=0.8\columnwidth]{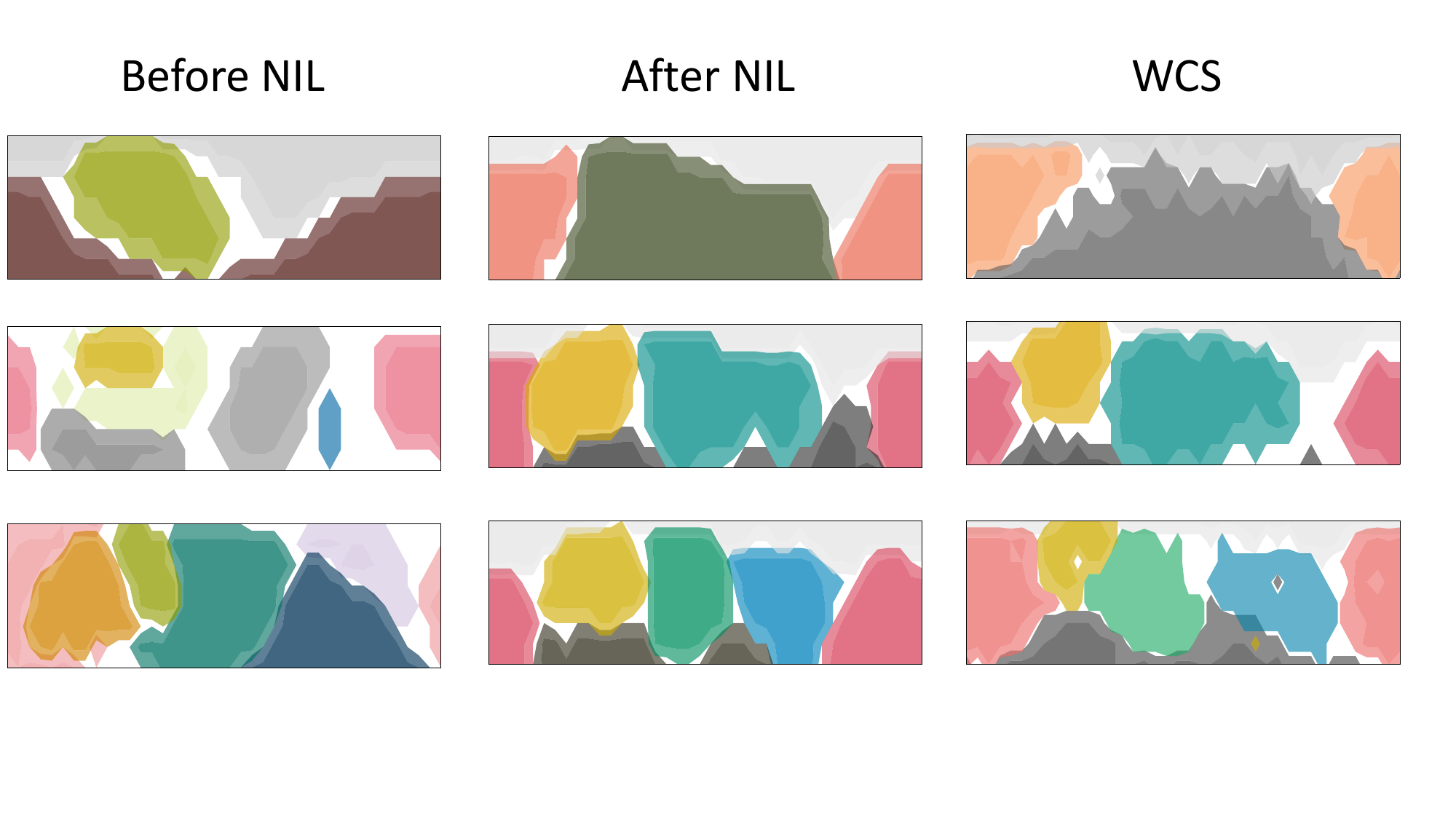}
    \caption{NIL transforms efficient color naming systems to become more similar to the WCS.  In each row, the left column shows an $\text{RM}_{\text{d}}$ system that was used to initialize NIL, the middle column shows the result of running NIL from that initialization state, and the right column shows a WCS system (from top to bottom: B\'et\'e, Colorado, Dyimini) that is similar to the NIL result.}
    \label{fig:il_maps}
\end{figure}

\begin{figure}[t]
    \centering
    \includegraphics[width=0.8\columnwidth]{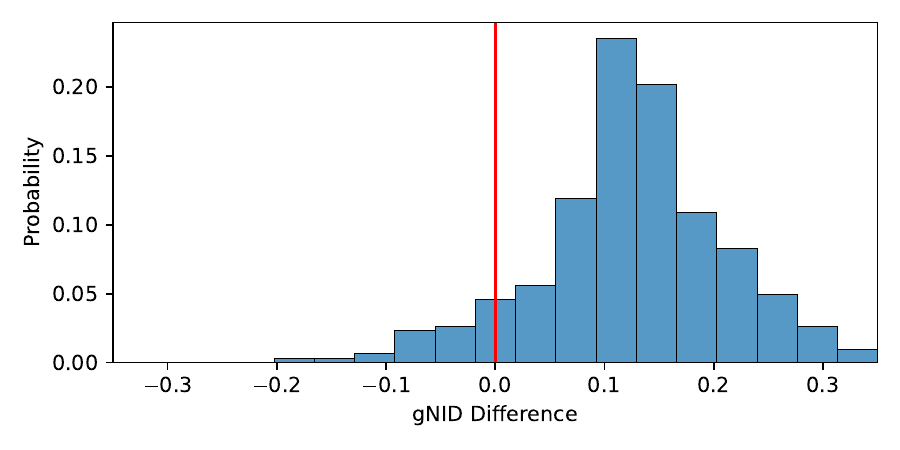}
    \caption{NIL tends to transform efficient $\text{RM}_{\text{d}}$ color naming systems to become more similar to the WCS.  The difference score is dissimilarity to WCS (minimum gNID to any WCS system) before NIL, minus the same quantity after NIL. Values above zero (marked by the vertical red line) indicate that NIL has brought a system closer to the systems of the WCS.  There is a clear trend towards positive values, indicating that NIL tends to transform already-efficient systems into systems that are more human-like.}
    \label{fig:il_dist}
\end{figure}

We also asked whether NIL would transform efficient systems that were dissimilar to those of the WCS (namely those of $\text{RM}_{\text{d}}$) into comparably efficient systems that were more similar to the WCS.  To test this, we initialized the NIL algorithm with a system sampled from $\text{RM}_{\text{d}}$, ran the NIL algorithm, and compared the initial system to the one that resulted from NIL.  Figure \ref{fig:il_maps} illustrates the beginning and end points of this process for a small set of systems, and shows that NIL transforms systems that are efficient but unlike the WCS into systems that are similar to particular WCS systems.

Figure \ref{fig:il_dist} shows the same general pattern but aggregated over all $\text{RM}_{\text{d}}$ systems.  For each NIL chain initialized with an $\text{RM}_{\text{d}}$ system, we measured the dissimilarity (gNID) of that initialized system to the most similar WCS system, and the gNID of the end result of NIL to its most similar WCS system.  It can be seen that NIL transforms $\text{RM}_{\text{d}}$ systems into systems that are more similar to the human systems of the WCS.  The mean gNID to WCS was $0.38$ before NIL and $0.25$ after, and the reduction in dissimilarity to WCS after applying NIL was significant (one-sided (paired) Wilcoxon signed-rank test, $n=302$, $T=1113$, $P \ll .001$). The median inefficiency of $\text{RM}_{\text{d}}$ is $\epsilon=0.09$ and the median inefficiency of the results of NIL is slightly lower at $\epsilon=0.07$, meaning that NIL made the already-efficient $\text{RM}_{\text{d}}$ systems slightly more efficient (one-sided (paired) Wilcoxon signed-rank test, $n=302$, $T=7716$, $P\ll.001$).  Thus, NIL moves already-efficient systems closer to the attested systems of the WCS, while maintaining and even slightly improving efficiency.  Finally, it is noteworthy that NIL with 3 terms converges to a system that is similar to a 3-term WCS system (see the top row of Figure \ref{fig:il_maps}), because 3-term systems are the one case in which IB optimal systems qualitatively diverge from human data  (\shortciteNP{Zaslavsky2018a}, p.\ 7941).  Thus, this is a case in which NIL appears to provide a better qualitative fit to the data than IB does. 


\subsection{Iterated learning alone, and communication alone}

So far, we have seen evidence that the NIL algorithm may provide a plausible model of the cultural evolutionary process by which human color naming systems become efficient.  We have referred to the result of the full NIL algorithm as IL+C systems, because these systems result from both iterated learning (IL) and communication (C).  This raises the question whether iterating learning alone, or communication alone, would yield comparable results.  

To find out, we ran two variants of the NIL algorithm.  One variant included only iterated learning but no communication (i.e.\ lines 6-8 of Algorithm \ref{alg:NIL} were omitted).  The other variant included communication but no iterated learning (i.e.\ there was only one pass through the main loop, which stopped at line 9); this is exactly the experiment that was performed by \shortciteA{Kageback2020}.  All other aspects of the algorithm were unchanged.  We refer to the results of the iterated-learning-only algorithm as IL (for iterated learning), and the results of the communication-only algorithm as C (for communication).

Comparison of the three panels of Figure \ref{fig:IL-vs-WCS} reveals that there are qualitative differences in the profiles of the systems produced by the 3 variants of the NIL algorithm (IL+C, IL, and C).  We have already seen that IL+C systems (top panel) are both efficient and similar to human systems; we also note that they lie within roughly the same complexity range as the human systems of the WCS.  In contrast, the IL systems (bottom left panel) skew toward lower complexity than is seen in human systems, and in fact about $6\%$ of the IL systems lie at the degenerate point $(0, 0)$ in the IB plane, at which there is a single category covering the entire color domain.  This skew toward simplicity is compatible with \shortciteauthor{Carr20}'s \citeyear{Carr20} claim that iterated learning provides a bias toward simplicity.
At the same time, the IL systems are not only simple but also quite efficient (i.e.\ informative for their level of complexity), which is in turn compatible with \shortciteauthor{Carstensen15}'s \citeyear{Carstensen15} claim that iterated learning provides some bias toward informativeness. 
Finally, the C systems (bottom right panel) show the opposite pattern: a bias toward higher informativeness, at the price of higher complexity, extending well above the complexity range observed in the human systems of the WCS.  

Taken together, these results suggest that iterated learning alone over-emphasizes simplicity, communication alone over-emphasizes informativeness, and iterated learning with communication provides a balance between the two that aligns reasonably well with what is observed in human color naming systems.  
Overall, these results suggest that iterated learning plus communication is a more plausible model of the cultural evolutionary process that leads to efficient human color naming systems than is either iterated learning alone, or communication alone, as these ideas are formalized in the NIL algorithm.

\subsection{The distribution of systems produced by IL+C} 

\begin{figure}[h]
    \centering
    \includegraphics[width=\textwidth]{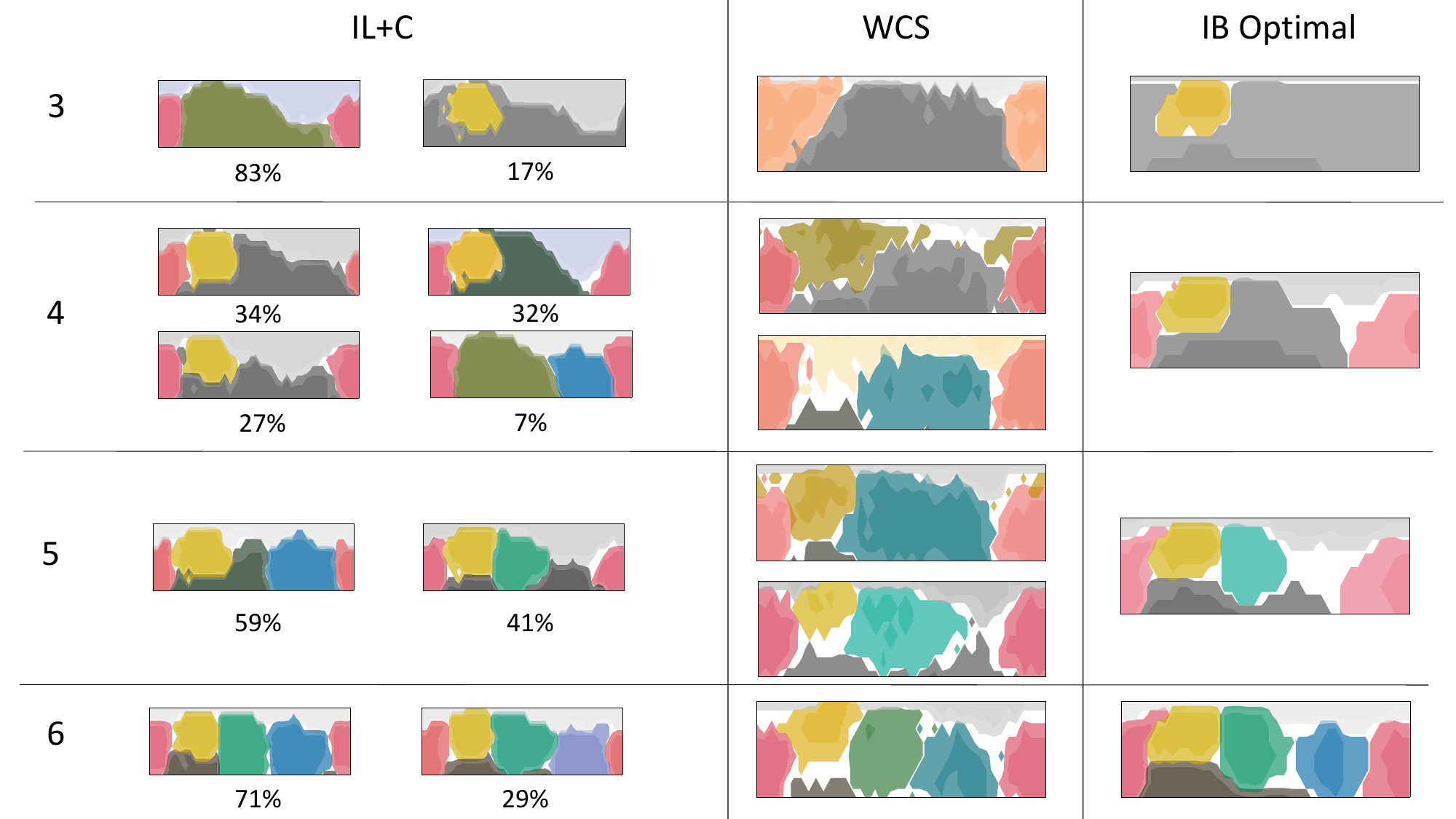}
    \caption{Representative IL+C systems (left column), WCS systems (middle column) and IB optimal systems (right column), with 3, 4, 5, and 6 color terms (rows). The $\%$ under each IL+C system indicates the percentage of IL+C systems in the corresponding cluster. The WCS systems are, from top to bottom: Nafaanra (iso: nfr, Ghana), Culina (iso: cul, Peru, Brazil), Waorani (iso: auc, Ecuador), Jicaque (iso: jic, Honduras), Berik (iso: bkl, Indonesia), and Kalam (iso: kmh, Papua New Guinea).}
    \label{fig:typ_systems_ilc}
\end{figure}
To further explore the distribution of systems produced by IL+C we grouped all IL+C systems from the main experiment based on the number of color terms, $K$, in the systems. For each number of color terms, we clustered the systems using  spectral clustering~\cite{Luxburg} with gNID as the dissimilarity measure. To find the appropriate number of clusters for each number of color terms, we performed spectral clustering with $C=2, 3, 4$ clusters and reported the clustering with the highest silhouette score~\cite{silscore} which is standard in clustering. Since spectral clustering does not return cluster centers, we take the system that minimizes the average pairwise gNID to all other systems in the cluster as a representative sample of that cluster. The resulting systems, for $K=3...6$, are presented in Figure~\ref{fig:typ_systems_ilc} along with some WCS systems and the optimal IB systems. The number under each representative IL+C system indicates the percentage of systems contained in the corresponding cluster. 

Interestingly, we see that the IL+C systems with three color terms appear in two clusters: a larger cluster that corresponds reasonably well to 3-term systems observed in the WCS, and a smaller cluster that is similar to the unattested IB optimal system. This suggests that there are two different optima that IL+C converges to: one human-like and the other corresponding to the IB optimal solution. The fact that the cluster corresponding to the IB solution is much smaller suggests that IL+C has a bias toward systems that are more similar to the WCS systems.  These results are compatible with the idea that the attested 3-term systems represent a local optimum that is easier to reach through a process of cultural evolution than is the IB optimal solution.

For the four term systems we observe that $93\%$ of the IL+C systems and up in clusters that corresponds well with the optimal IB system and one of the WCS systems shown in Figure~\ref{fig:typ_systems_ilc}. The last $7\%$ of the systems end up in a cluster where the representative system does not have a clear dark term but instead a blue and green term which is a combination not found in the WCS data. Moreover, for both $K=5$ and $K=6$ we observe that all IL+C clusters seem to correspond fairly well with systems in WCS and the IB optimal systems.

\subsection{Learnability and convexity}

An influential idea holds that human categories form convex regions in a given conceptual space \shortcite{Gardenfors00}.  In the case of color, a natural space for testing this claim is CIELAB space (Figure \ref{fig:maps}, top right panel), and \citeA{Jager10} has in fact shown that the natural color categories found in the WCS are convex sets in CIELAB space --- supporting the convexity claim of \citeA{Gardenfors00} in the domain of color.  More recently, \citeA{Steinert-Threlkeld20} have extended this line of thought by arguing that convex color categories are easier to learn than are non-convex ones, and that this greater learnability helps to explain why human color categories tend to be convex.

We sought to situate this argument relative to the one we have been advancing here.  Intuitively, it seems plausible that the artificial RM systems we have considered above should also be convex, because they are based on  spherical Gaussian-shaped kernels --- but as we have seen, many of these RM systems are quite dissimilar to the human systems of the WCS.  This suggests that convexity may be a necessary but not sufficient criterion for characterizing human-like semantic categories, a suggestion with which proponents of the convexity argument are comfortable (P.\ G\"ardenfors, G. J\"ager, personal communication).  To probe this possibility further, we assessed the convexity, the (non-iterated) learnability, and the efficiency of the WCS systems, the RM systems, and an additional set of baseline systems that draw category distinctions based only on hue.  These hue-based systems were designed to be convex but not similar to human systems.  Specifically, for vocabulary sizes $K=3...10$ we divided the Munsell chart into equally sized categories by grouping together color chips based on their hue only; in case equally sized categories were not possible we created $K-1$ equally sized categories and added the remaining color chips to the last category.  Example hue-based systems are shown in Figure~\ref{fig:hue-systems}:
\begin{figure}[b]
    \centering
    \includegraphics[width=0.3\textwidth]{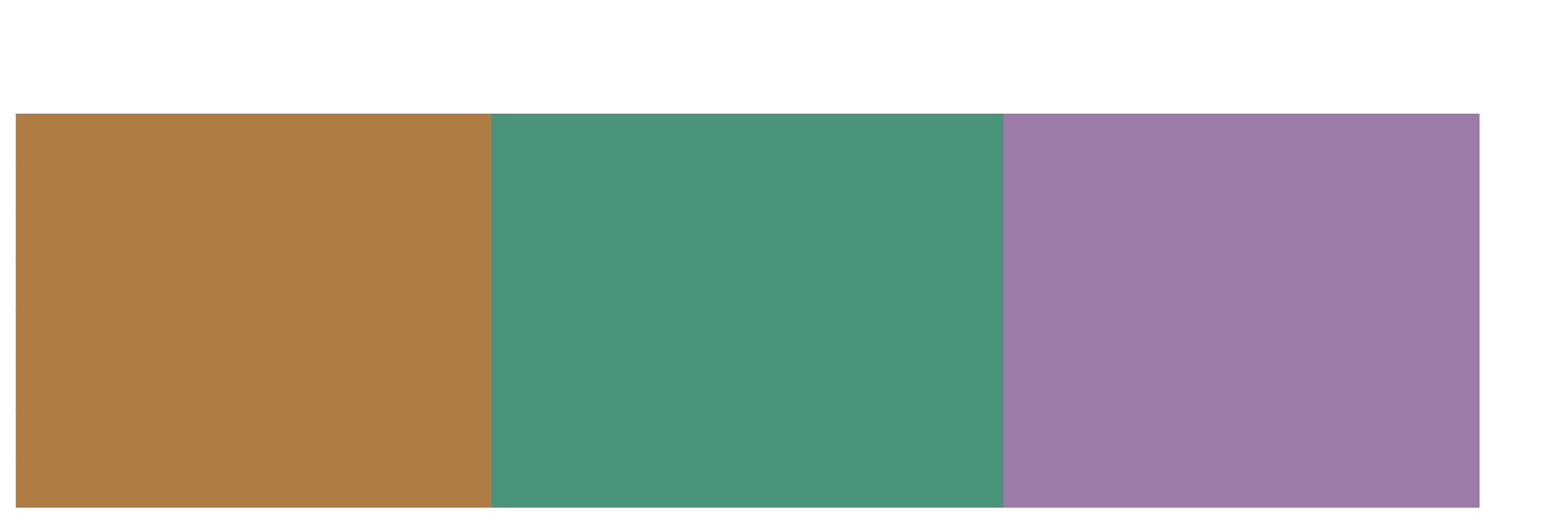} 
    \includegraphics[width=0.3\textwidth]{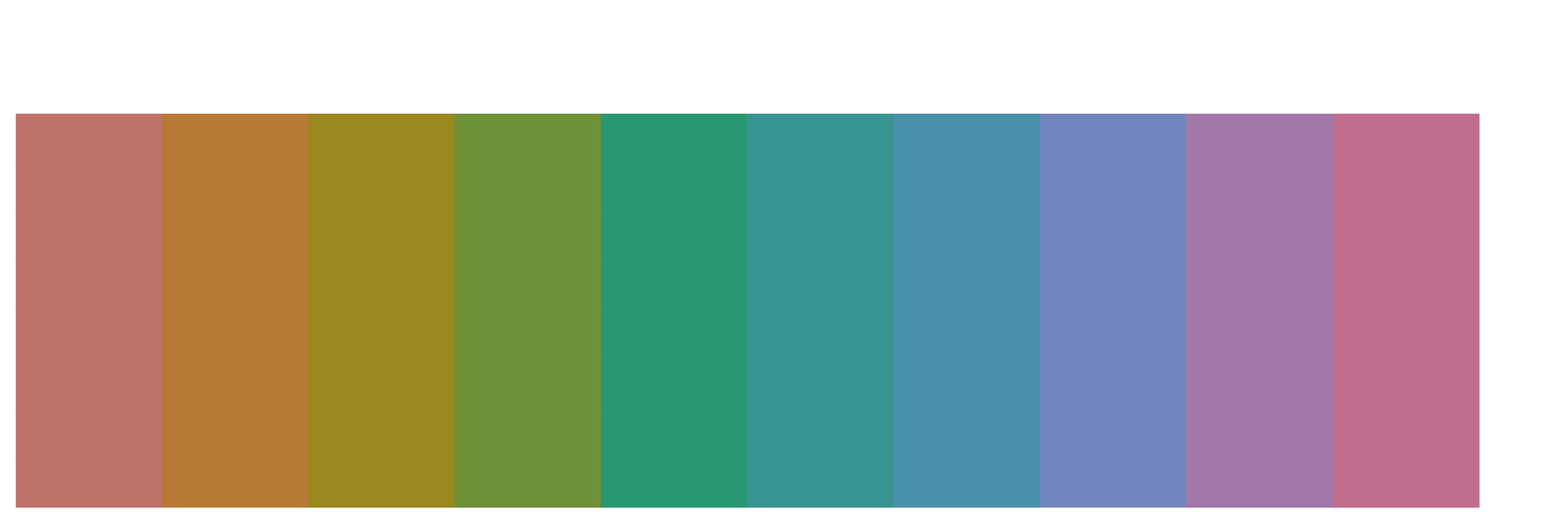} 
    \caption{Hue-based artificial systems, with 3 (left) and 10 (right) categories.}
    \label{fig:hue-systems}
\end{figure}
these are deterministic systems in which hue column fully determines the category to which a given chip belongs.

To assess the convexity of a color naming system, we adopted the measure of \citeA{Steinert-Threlkeld20}.  They took the degree of convexity of a single category, named by a word $w$, to be:

\begin{align*}
    \text{dcc}(w) := \frac{|w|}{|\text{ConvHull}(w)|}
\end{align*}

\noindent 
where $|\cdot|$ is the size of a set, i.e.\ the number of color chips in that set, and $\text{ConvHull}(w)$ is the convex hull, in CIELAB space, of those chips in category $w$.  Thus, $\text{dcc}(w)$ gives us the proportion of those chips in the convex hull of category $w$ that are also in the category $w$ itself.  For a perfectly convex category, this proportion will be $1$.  \citeA{Steinert-Threlkeld20} then defined the degree of convexity of an entire system $S$ of categories to be the average, weighted by category size, of $\text{dcc}(w)$ across categories $w$ in $S$:

\begin{align*}
    \text{dc}(S) := \frac{\sum_{w \in S} |w| \cdot \text{dcc}(w)}{\sum_{w \in S}|w|}
\end{align*}

\noindent
A $\text{dc}(S)$ value of $1$ corresponds to a system of perfectly convex color categories.\footnote{This method assumes deterministic rather than probabilistic category membership.  When applying this method to probabilistic systems, we first converted the probabilistic system to a deterministic one by assigning each chip to the modal category for that chip;  we then applied this convexity measure to the resulting deterministic system.}

To assess the (non-iterated) learnability of a color naming system, we took a system to be easily learned to the extent that a neural network learner {\it generalizes} the system well --- that is, to the extent that the learned system matches the one from which training data was sampled.  We assessed this by considering only the learning phase of the NIL algorithm, and considering only the speaker's learning (specifically lines 3-5 of Algorithm \ref{alg:NIL}), leaving all parameters unchanged.  We then measured the gNID between the learned system and the system from which training data was drawn.  During training, the agent sees only part of the entire system, so this gNID is a measure of how well the agent generalizes from the data it receives.  To mitigate possible effects caused by sampling the training dataset, we performed each experiment over $10$ independent runs and averaged.

\begin{figure}[t]
\centering
\includegraphics[width=.45\linewidth]{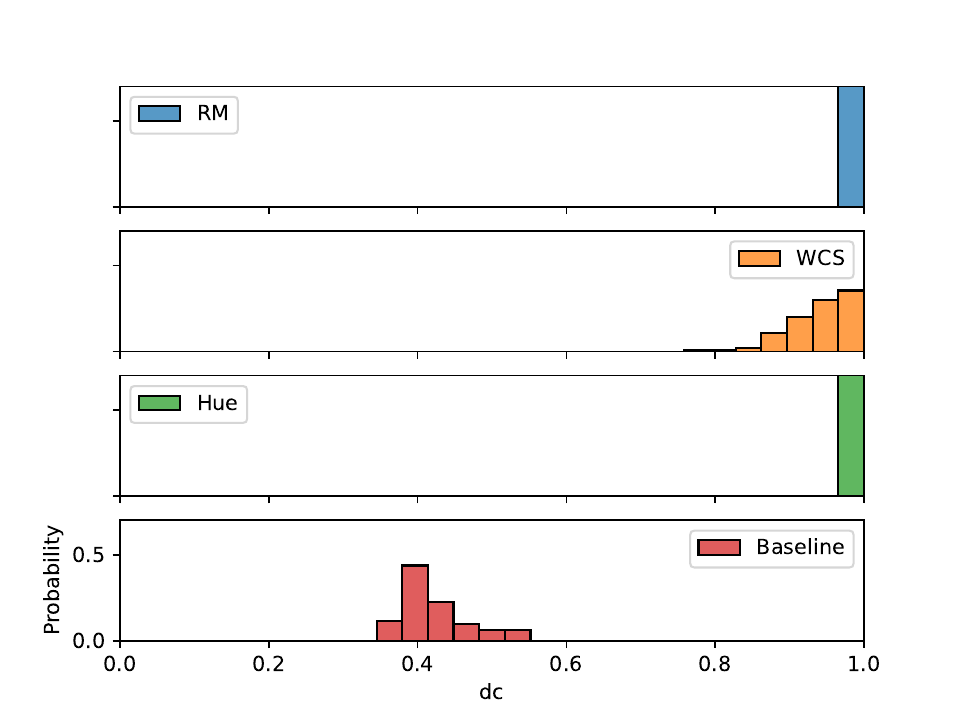}
\includegraphics[width=.45\linewidth]{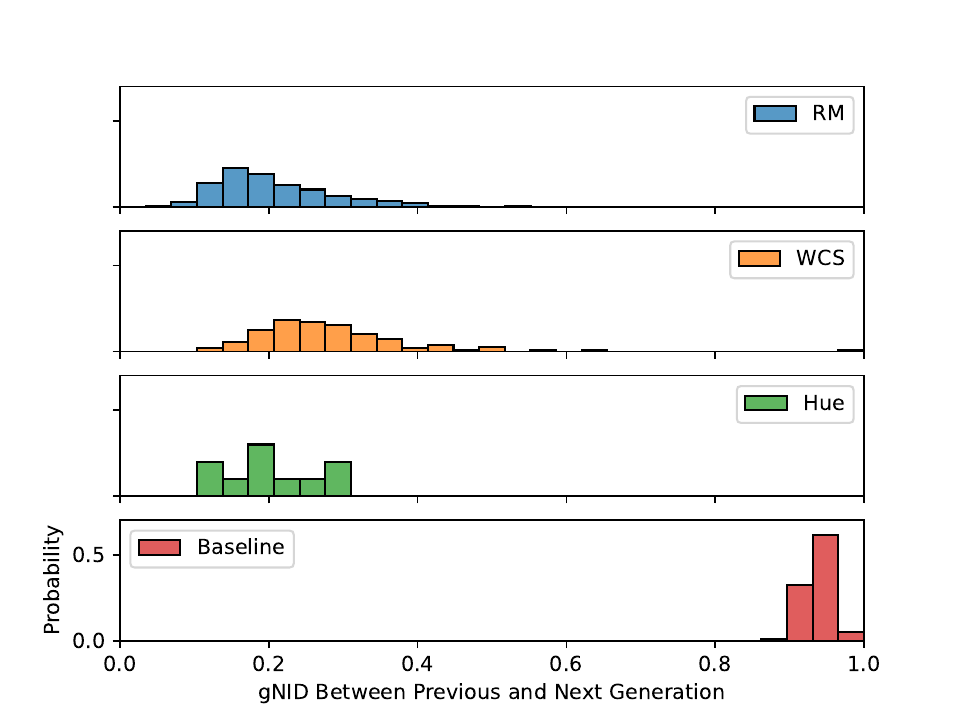}
\caption{{\bf Left panel: Convexity.}  Convexity for different types of category systems.  The natural systems of the WCS, and the artificial RM and hue-based systems, are all highly convex when compared with a baseline of randomly generated systems in which each color chip is assigned to a category selected uniformly at random (labeled ``Baseline''). We generated such random systems with $k=3...10$ color categories and for each $k$ we drew $10$ random systems.  {\bf Right panel: Learnability.} 
 Ease of learning is assessed by how well a learner generalizes, and generalization is measured by gNID between a learned system and the system from which training data was drawn.  Artificial RM and hue-based systems show  generalization that is no worse than that of natural WCS systems.}
\label{fig:convexity-learnability}
\end{figure}

\begin{figure}[t]
    \centering
    \includegraphics[width=0.8\textwidth]{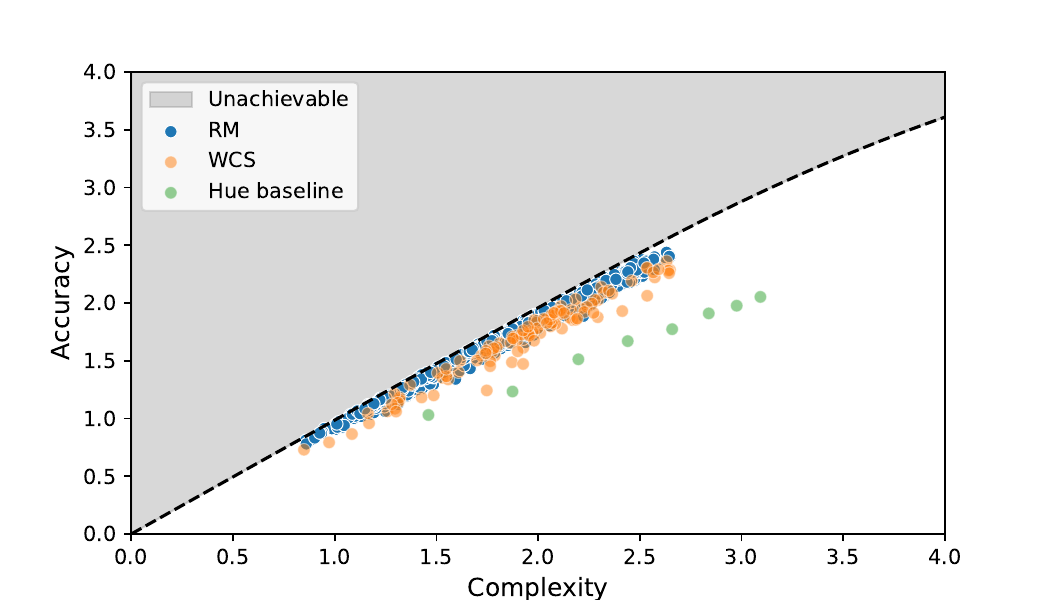}
    \caption{Some convex and learnable category systems are not efficient. Efficiency of the artificial hue-based systems (green dots), compared with that of the artifical RM (blue dots) and natural WCS (orange dots) systems.}
    \label{fig:hue_baseline}
\end{figure}

We assessed the convexity, the learnability, and the IB efficiency of the (natural) WCS, (artificial) RM, and (artificial) hue-based systems.  Convexity results are shown in Figure~\ref{fig:convexity-learnability} (left panel), and learnability results are shown in Figure~\ref{fig:convexity-learnability} (right panel). 
All three types of system are highly convex, with the artificial RM and hue-based systems being slightly more convex than the natural WCS systems -- perhaps because the natural systems include noise.
Moreover, in line with the expectation that convex systems will be learnable, all three types of system show good generalization, with no advantage for the natural WCS systems over the artificial RM and hue-based systems.  These results confirm that convex systems tend to be highly learnable, and also highlight that something beyond convexity and (non-iterated) learnability must play a role in differentiating human systems from artificial semantic systems that do not resemble them.  Finally, Figure~\ref{fig:hue_baseline} shows that artificial hue-based systems are not especially efficient --- in contrast with artificial RM systems and natural WCS systems.  We take these results to suggest that convexity and learnability are only a partial answer to the question of what characterizes human semantic categories --- and that a fuller answer may be provided by iterated learning and communication operating together, as a model of cultural evolution that leads toward efficient and human-like systems of semantic categories. 

\balance

\section{Discussion}

We have shown (1) that there exists a reasonably sized class of color naming systems that are highly efficient in the IB sense but dissimilar from human systems; (2) that iterated learning plus communication, as captured in the NIL algorithm, leads to color naming systems that are both efficient in the IB sense and similar to human systems, and (3) that iterated learning alone, communication alone, and convexity alone, do not yield that result as clearly.  These findings help to answer some questions, and also open up others.

As we have noted, the existence of highly efficient systems that do not align with human ones is not in itself surprising.  IB is a non-convex optimization problem \shortcite{Tishby-1999,Zaslavsky2018a}, so multiple optima and near-optima are to be expected.  However we feel that our identification of such systems may nonetheless be helpful, because it highlights just how many such systems exist, and just how dissimilar from human systems they sometimes are --- which helps to make sense of \shortciteauthor{Chaabouni}'s
\citeyear{Chaabouni} finding that simulations of cultural evolution can lead to color naming systems that exhibit high IB efficiency but deviate to some extent from human systems.  This in turn highlights the importance of identifying cultural evolutionary processes that avoid these local near-optima and instead converge toward systems we find in human languages. 

We have argued that iterated learning plus communication, as cast in the NIL algorithm, is such a process, and that it provides a better account than either iterated learning alone, or communication alone.  This idea, and our findings supporting it, may help to resolve a tension in the literature.  As we have noted, \shortciteA{Carstensen15} argued that iterated learning alone can lead to informative semantic systems, whereas \shortciteA{Carr20} argued that iterated learning provides a bias for simplicity, and communication provides a bias for informativeness (see also \shortciteNP{kirby-et-al-2015} for a similar argument concerning linguistic form).  Our finding that both forces are needed to account for the data aligns with \shortciteauthor{Carr20}'s \citeyear{Carr20} claim.  However our finding that learning alone also converges to efficient and thus informative systems -- although often to overly simple ones -- helps to make sense of \shortciteauthor{Carstensen15}'s \citeyear{Carstensen15} findings.

\shortciteauthor{tucker-et-al-2022} \citeyear{tucker-et-al-2022} explored a model in which agents communicated while optimizing an objective based on the IB objective, and found that this model produced systems matching the human systems in the WCS.
Our results suggest that the explicit penalty against complexity that was incorporated in their objective, as in the IB objective, may not be necessary to explain the WCS data, and that cultural transmission, here modelled using iterated learning, naturally produces a bias towards simplicity \shortcite{Carr20}.

It is natural to think of NIL, or any such process of cultural evolution, as a means by which the abstract computational goal of optimal efficiency might be attained or approximated.  The optimally efficient color naming systems on the IB curve closely resemble those in human languages \shortcite{Zaslavsky2018a}, and the IL+C systems are likewise highly efficient and similar to those in human languages.  However, there is an important exception to this pattern.  As noted above, in the case of 3-term systems, the IB optimal system qualitatively differs from the color naming patterns found in the WCS (\shortciteNP{Zaslavsky2018a}, p.\ 7941), whereas IL+C systems often qualitatively match attested 3-term systems (recall the top rows of Figures \ref{fig:il_maps} and \ref{fig:typ_systems_ilc}).  Thus, in this one case, it appears that human languages do not attain the optimal solution or something similar to it, and instead attain a somewhat different near-optimal solution that is apparently more easily reached by a process of cultural evolution.

A major question left open by our findings is exactly why we obtain the results we do.  NIL is just one possible evolutionary process, and we have seen that that process accounts for existing data reasonably well.  It makes sense intuitively that NIL strikes a balance between the simplicity bias of iterated learning and the informativeness bias of communication \shortcite{Carr20,kirby-et-al-2015} --- but what is still missing is a finer-grained sense for exactly which features of this detailed process are critical, vs.\ replaceable by others, and what the broader class of such processes is that would account well for the data (e.g.\ \shortciteNP{tucker-et-al-2022}).  A related direction for future research concerns the fact that the evolutionary process we have explored is somewhat abstract and idealized, in that agents communicate with little context or pragmatic inference.  Actual linguistic communication is highly context-dependent, and supported by rich pragmatic inference --- it seems important to understand whether our results would still hold in a more realistic and richer environment for learning and interaction.  Finally, we have focused here on the domain of color, but the ideas we have pursued are not specific to color, so another open question is the extent to which our results generalize to other semantic domains.  

\nocite{Hunter:2007}
\nocite{Waskom2021}
\nocite{scikit-learn}
\nocite{paszke2019pytorch}
\nocite{mckinney2010data}
\nocite{scipy}

\balance

\section*{Acknowledgments}

An earlier version of this paper appeared in the Proceedings of the 45th Annual Meeting of the Cognitive Science Society (2023).  We thank Noga Zaslavsky and 3 anonymous reviewers for helpful comments on that earlier version of the paper. Any errors are our own.  Author contributions: EC, DD, and TR designed the research; EC performed the research; EC analyzed the data; and EC, DD, and TR wrote the paper.
EC was funded by Chalmers AI Research (CHAIR) and the Sweden-America Foundation (SweAm). The computations were enabled by resources provided by the National Academic Infrastructure for Supercomputing in Sweden (NAISS) partially funded by the Swedish Research Council through grant agreement no. 2022-06725.

\setlength{\bibleftmargin}{.125in}
\setlength{\bibindent}{-\bibleftmargin}

\bibliography{library}

\clearpage

\appendix

\section{The framework of Zaslavsky et al.\ (2018)}\label{appendix:IB}

\begin{figure}[t]
    \centering
    \includegraphics[width=0.5\textwidth]{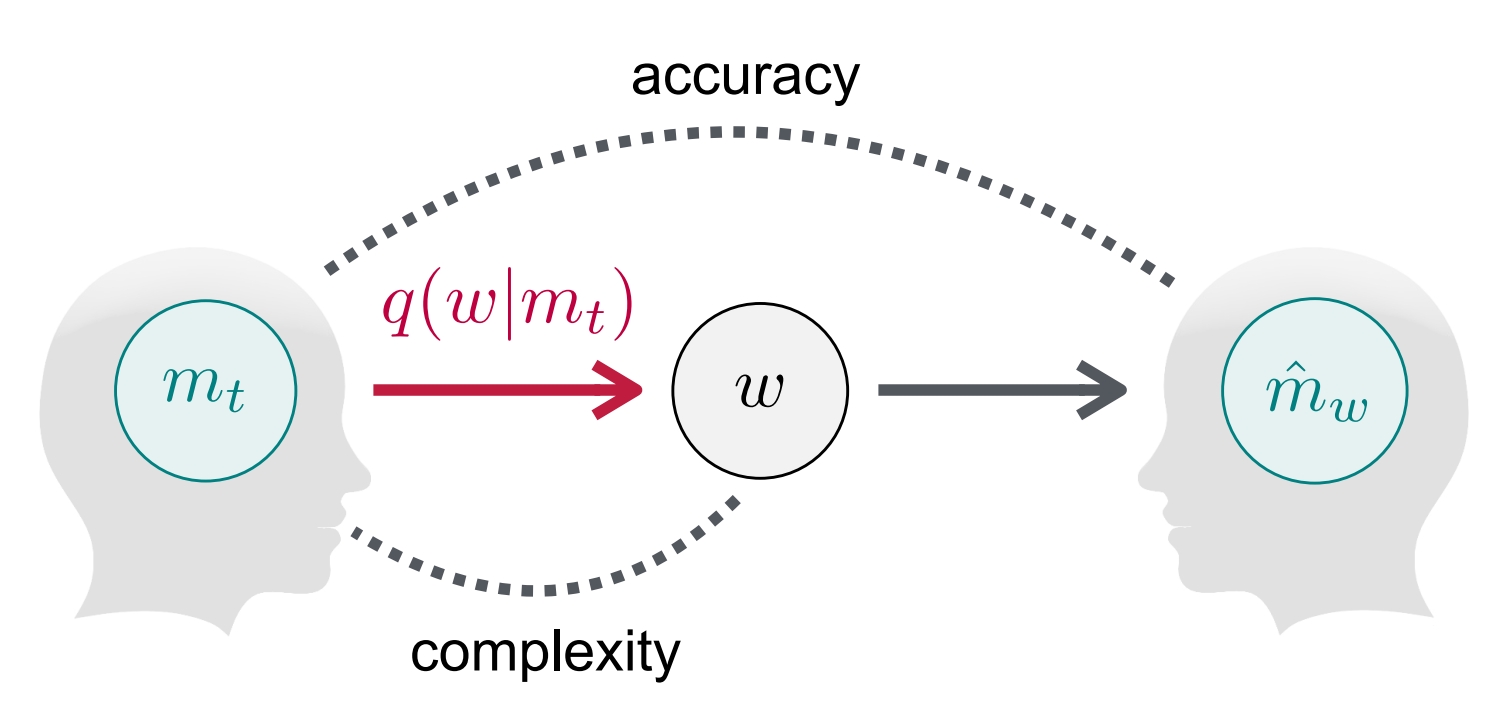}
    \caption{The framework of Zaslavsky et al.\ (2018). 
    A speaker communicates a specific referent to a listener by producing a word.  The IB principle provides formal specifications of various quantities associated with this communicative act; see text for details.  The figure is from Zaslavsky et al. (2021).}
    \label{fig:comm_model}
\end{figure}

\shortciteA{Zaslavsky2018a} cast the notion of  efficiency in terms of an independent information-theoretic principle, the Information Bottleneck (IB) principle~\shortcite{Tishby-1999}.  In the framework of \shortciteA{Zaslavsky2018a}, a semantic system is considered efficient to the extent that it achieves an optimal tradeoff between the complexity of a system, and the accuracy of communication that that system supports.  These notions are grounded in the communicative scenario illustrated in Figure~\ref{fig:comm_model}, in which a speaker attempts to communicate with a listener about referents in a given domain universe $U$, in our case the domain of color.  Here, the speaker considers a specific target color $t \in U$ and holds it in mind in the form of a mental representation $m_t$, which is a probability distribution over color space (CIELAB; recall Figure~\ref{fig:maps}), centered at $t$.    To communicate that mental representation, the speaker utters a word $w$, drawn from a language-specific probabilistic encoder $q(w|m_t)$ that maps from meanings $m_t$ to words $w$; this encoder $q(w|m_t)$ is the semantic system by which the speaker and listener communicate.  The listener then produces, on the basis of the uttered word $w$, a mental representation $\hat{m}_w$ that is the listener's reconstruction of the speaker's original representation $m_t$.  Casting this simple communicative scenario in terms of the IB principle results in formal definitions of four quantities that are central to the IB formalization of efficiency, and on which we rely in our work: {\em complexity, accuracy,} $\epsilon$, and {\em gNID}.

The complexity of a semantic system $q$ is given by $I_q(M_t;W)$, i.e.\ the mutual information between the speaker's mental representation $m_t$ and the word $w$ used to express it.  The greater the complexity of the system, the more information the word $w$ carries about the speaker's mental representation $m_t$.  The accuracy of a semantic system is given by $I_q(W;U)$, which can be shown to capture the similarity of the speaker's and listener's mental representations (see \shortciteNP[p.\ 7939]{Zaslavsky2018a}).  The core idea of efficiency in this framework is to obtain the greatest accuracy possible for a given level of complexity --- i.e.\ to communicate as precisely as possible for a given amount of information sent.  An optimally efficient semantic system $q$ is thus one that minimizes the IB objective function:
\begin{align*}
  \mathcal{F}_\beta [q] = I_q(M_t;W) - \beta I_q(W;U)
\end{align*}
where $\beta \ge 0$ is a tradeoff parameter that controls the relative weight given to complexity and accuracy.  Those systems $q^*$ that minimize this objective function for different values of $\beta$ yield the IB theoretical limit of
efficiency; that is, these are the systems with the greatest possible accuracy for each level of complexity.  \shortciteA{Zaslavsky2018a} showed that human color naming systems achieve near-optimal efficiency in the IB sense, and that fully IB-optimal systems often closely correspond to color naming systems in human languages.

In our analyses, we also make use of two other quantities from the framework of \shortciteA{Zaslavsky2018a}. 
 First, $\epsilon$ measures the inefficiency of a semantic system, or its deviation from optimal efficiency, as described on p.\ 7939 of their article.
Finally, we follow \shortciteA{Zaslavsky2018a} in using their gNID measure to measure the dissimilarity between two semantic systems, as described on p.\ 7942 of their article.

\end{document}